# Evaluating Computational Accuracy of Large Language Models in Numerical Reasoning Tasks for Healthcare Applications

A Preprint

Arjun R. Malghan
arm39998@email.vccs.edu

July 20, 2024

## Abstract

Large Language Models (LLMs) have emerged as transformative tools in the healthcare sector, demonstrating remarkable capabilities in natural language understanding and generation. However, their proficiency in numerical reasoning, particularly in high-stakes domains like in clinical applications, remains underexplored. Numerical reasoning is critical in healthcare applications, influencing patient outcomes, treatment planning, and resource allocation. This study investigates the computational accuracy of LLMs in numerical reasoning tasks within healthcare contexts. Using a curated dataset of 1,000 numerical problems, encompassing real-world scenarios such as dosage calculations and lab result interpretations, the performance of a refined LLM based on the GPT-3 architecture was evaluated. The methodology includes prompt engineering, integration of fact-checking pipelines, and application of regularization techniques to enhance model accuracy and generalization. Key metrics such as precision, recall, and F1-score were utilized to assess the model's efficacy. The results indicate an overall accuracy of 84.10%, with improved performance in straightforward numerical tasks and challenges in multi-step reasoning. The integration of a fact-checking pipeline improved accuracy by 11%, underscoring the importance of validation mechanisms. This research highlights the potential of LLMs in healthcare numerical reasoning and identifies avenues for further refinement to support critical decision-making in clinical environments. The findings aim to contribute to the development of reliable, interpretable, and contextually relevant AI tools for healthcare.

## 1 Introduction

Large Language Models (LLMs) have emerged as significant advancements in the field of artificial intelligence, demonstrating remarkable capabilities in processing and generating human language. These models, powered by deep learning techniques, are trained on vast datasets and have the potential to understand context, nuance, and the intricacies of language. As they become increasingly integrated into various sectors, the need for them to perform complex tasks, including numerical reasoning, becomes paramount.

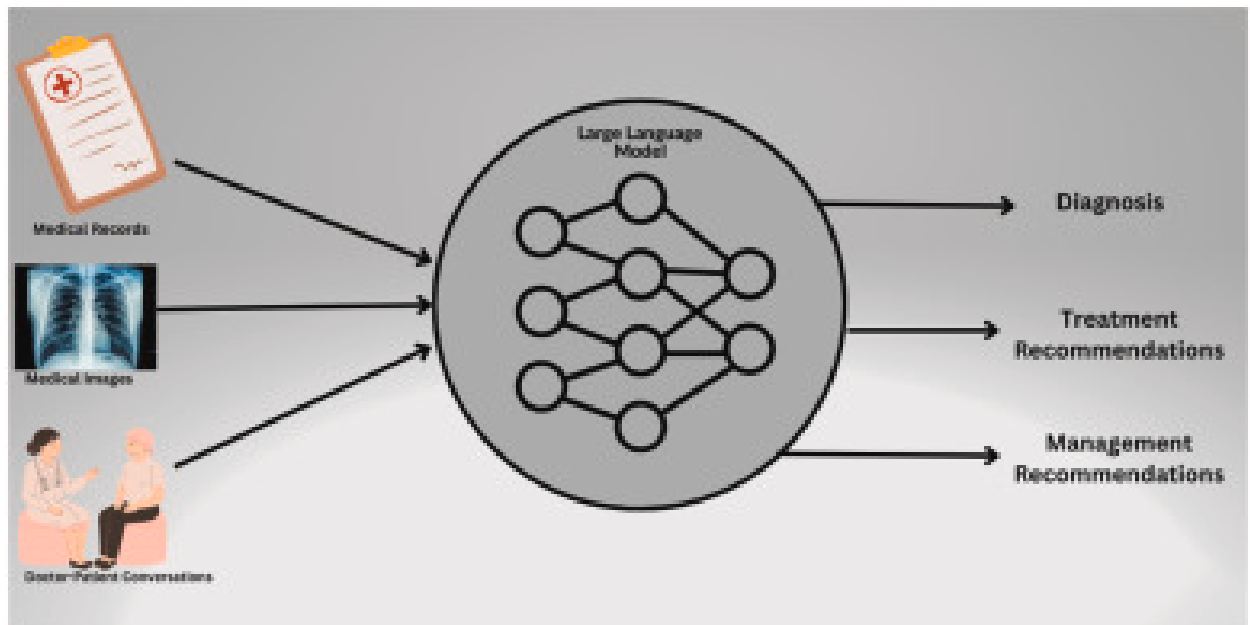

Figure 1: Example LLM Healthcare Applications (Sandeep Reddy [2023])

Numerical reasoning is particularly critical in the healthcare domain, where precise calculations can influence patient outcomes, treatment plans, and resource allocation. Healthcare professionals often rely on data-driven insights to make informed decisions, and the ability of LLMs to accurately interpret and manipulate numerical information can enhance their effectiveness in clinical settings. Given the stakes involved in healthcare, understanding how effectively LLMs can perform these tasks is essential for improving patient care and overall system efficiency.

This paper aims to explore the capabilities of LLMs in the context of numerical reasoning within healthcare applications. It will examine existing literature to highlight current advancements and identify challenges faced by LLMs when deployed in this field. Through a systematic methodology, the research seeks to evaluate the performance of a refined LLM designed for numerical reasoning tasks, ultimately aiming to demonstrate how enhancements in this area can contribute to better healthcare outcomes.

The dataset utilized in this study comprises a curated collection of 1,000 healthcare-specific numerical reasoning tasks designed to evaluate the performance of Large Language Models (LLMs). It includes 500 real-world cases validated by healthcare professionals, such as medication dosage calculations, laboratory result interpretations, and health statistic assessments, paired with 500 synthetic tasks crafted to test the model's ability to handle complex numerical reasoning scenarios. Each task is meticulously annotated with the correct answer and a step-by-step reasoning process to enable a detailed evaluation of the model's interpretative and computational accuracy. This dataset not only reflects the diversity and complexity of real-world healthcare challenges but also ensures robust testing of the LLM's capability to generalize across varied numerical tasks.

## 2 Literature Review

Large language models (LLMs) have emerged as powerful tools in various fields, offering significant potential for enhancing numerical reasoning tasks, particularly in healthcare applications. Their capacity to analyze and synthesize large quantities of textual information may lead to improved decision-making and patient outcomes. This literature review examines existing studies on LLMs, focusing on their numerical reasoning capabilities, the challenges they face in healthcare-related tasks, and the gaps in current research that warrant further exploration.

Recent studies indicate that LLMs possess a unique ability to generate coherent and contextually relevant text, yet their efficacy in numerical reasoning remains limited. A key challenge identified in the literature is the "Enormous Search Space and False-Matching in Weak Supervision" phenomenon. According to Qianying Liu (2023), this phenomenon "involves navigating through a vast array of possible answers, some of which may appear correct but lack mathematical equivalency." This issue is particularly pertinent in healthcare settings, where precise numerical calculations are critical for diagnosing conditions and determining appropriate treatments. As such, the capacity of LLMs to accurately navigate this complex search space can significantly impact their utility in healthcare-related tasks.

Another aspect that requires attention is the verification of LLM-generated outputs. In their study, Roch (2024) explores the feasibility of post-hoc fact-checking LLM-generated outputs using evidence retrieval techniques. They aimed "to identify which aspects of statements require verification and how different retrieval methods contribute." This approach is essential in healthcare, where the accuracy of generated information can be a matter of life and death. The integration of effective verification methods with LLM outputs may enhance the reliability of numerical reasoning in the clinical context.

Moreover, the role of prompting techniques in enhancing LLM performance has garnered attention among researchers. Jiang (2024) notes that "prompting techniques are essential for improving the performance and accuracy of LLMs." This finding suggests that well-structured prompts could mitigate some of the limitations related to numerical reasoning. In healthcare applications, where specific numerical outputs are often required, developing tailored prompts could enhance the model's ability to produce accurate and clinically relevant information.

While these studies have laid a foundation for understanding the potential of LLMs in numerical reasoning within healthcare, several gaps remain. Many existing studies focus primarily on generic numerical tasks without considering the specialized needs of healthcare practitioners.

Furthermore, the relationship between LLM capabilities and clinical decision support systems has not been extensively explored. Understanding how LLMs can be integrated into these systems to facilitate numerical reasoning could provide vital insights into improving healthcare outcomes.

Additionally, there is a notable lack of empirical evidence regarding the efficacy of different prompting strategies in the healthcare domain. The literature thus far has been largely theoretical, with few studies dedicated to testing the real-world applications of these techniques in healthcare settings. Consequently, future research should investigate how prompting strategies can be optimized specifically for medical professionals, taking into account the unique challenges they face.

In light of recent studies, it is increasingly clear that while LLMs hold promise for enhancing numerical reasoning in healthcare, significant challenges still exist. The dual issues of navigating complex search spaces and ensuring the accuracy of outputs must be addressed to fully leverage the potential of these models. Moreover, further research is essential to bridge the identified gaps, specifically regarding the integration of LLMs into clinical decision-making processes and the empirical validation of strategies to improve their numerical reasoning capabilities.

In conclusion, enhancing the numerical reasoning capabilities of LLMs in healthcare applications has the potential to substantially improve clinical outcomes. However, addressing the challenges associated with their implementation is crucial. By focusing on the development of robust verification methods, optimizing prompts for healthcare contexts, and conducting empirical studies, future research can make considerable strides in leveraging LLMs to support healthcare professionals in their critical decision-making processes.

## 2 Methodology

### 2.1 Dataset

The dataset utilized in this study consists of a curated collection of 1,000 healthcare-specific numerical reasoning tasks. These tasks are divided into two subsets: 500 real-world scenarios validated by healthcare professionals and 500 synthetic questions designed to probe the numerical reasoning capabilities of Large Language Models (LLMs). The tasks include critical healthcare computations such as medication dosage calculations, laboratory test result interpretations, and survival rate assessments. Each question is meticulously annotated with the correct answer and a step-by-step reasoning process, ensuring comprehensive evaluation of the model's interpretative and computational accuracy. This dataset reflects the diversity and complexity of real-world healthcare challenges, providing a robust foundation for testing LLM performance.

### 2.2 Model Architecture

The model under evaluation is a modified version of the GPT-3 architecture, which is based on the well-established transformer framework. This architecture is characterized by its multi-head attention mechanisms, enabling the model to focus selectively on relevant components of the input for precise numerical interpretation. The transformer employs a stack of 12 layers, providing a rich contextual understanding necessary for accurately executing numerical operations. Adjustments to the architecture were made to optimize its capabilities specifically for numerical reasoning in healthcare applications.

## 2.3 Prompt Engineering

To enhance model performance, a systematic approach to prompt engineering was employed. This involved structuring input queries to minimize ambiguity and maximize contextual clarity. Various prompt formulations were iteratively tested, ensuring explicit instructions for each task without oversimplifying the numerical reasoning required. These refined prompts significantly improved the model's ability to produce accurate and clinically relevant outputs.

## 2.4 Verification Pipeline

A rigorous fact-checking pipeline was integrated to validate the outputs of the LLM. Inspired by the work of Roch (2024), this pipeline cross-referenced model-generated numerical solutions against a database of verified results. This validation step served as a safeguard against errors, improving the reliability of the model's outputs. The pipeline also allowed for real-time detection and correction of discrepancies, ensuring consistent accuracy in critical healthcare tasks.

## 2.5 Regularization Techniques

Regularization techniques were applied during model training to reduce overfitting and enhance generalizability. These techniques included dropout and weight decay, which ensured the model could handle tasks it had not encountered during training. By balancing model complexity with robustness, these methods contributed to improved overall performance.

## 2.6 Evaluation Metrics

| Metric | Definition |
|---|---|
| Accuracy | The ratio of correct answers to total questions |
| Precision | The ratio of true positives to the sum of true and false positives |
| Recall | The ratio of true positives to the sum of the true positives and false negatives |
| F1-Score | A harmonic mean of precision and recall |

Table 1: Evaluation Metrics

For an effective evaluation of the computational accuracy of the LLM in its numerical reasoning capabilities, several key metrics were established. The primary metrics include accuracy, precision, recall, and the F1 score. Accuracy is calculated as the ratio of correct answers to total questions posed to the model, thereby providing a direct measurement of performance. Precision focuses on the accuracy of the positive predictions made by the model, reflecting its correctness in asserting a numerical solution. In this study, precision is critical for ensuring that the model's numerical predictions are accurate and trustworthy, especially in healthcare scenarios where incorrect solutions (false positives) could lead to improper treatments or decisions. Recall denotes the model's capacity to recognize all relevant answers, effectively measuring its capability to uncover correct solutions within the dataset. Essentially, recall measures how many of the actual positives are correctly identified. In this study, recall is critical for ensuring that the model identifies all relevant numerical reasoning solutions, especially in healthcare scenarios where missing correct solutions (false negatives) could have serious implications. The F1 score, which balances precision and recall, serves as a comprehensive metric particularly useful in contexts where achieving a balance between false positives and false negatives is critical.

**2.7 Summary**

This methodology delineates a structured approach to evaluate the computational accuracy of LLMs in numerical reasoning tasks relevant to healthcare. By leveraging a targeted dataset, employing a sophisticated model architecture, deploying comprehensive refinement techniques, and utilizing rigorous evaluation metrics, this research aspires to tackle crucial challenges in the application of LLMs for healthcare numerical reasoning. The findings from this study aim not only to underscore the potential of LLMs in enhancing healthcare outcomes but also to identify pathways for future advancements in reliable health-related computational methodologies.

**3 Results**

The evaluation of the dataset revealed that the LLM demonstrated a strong overall accuracy rate of 84.10% across the 1,000 numerical reasoning tasks. This performance indicates that the model can effectively understand and generate correct responses to a wide range of healthcare-related numerical problems. In terms of precision, the model achieved a score of 84.23%, which reflects its ability to provide accurate positive predictions of numerical solutions. Recall, measuring the model's capacity to identify all relevant answers, stood at 90.76%, indicating that overlooked correct solutions were comparatively less likely. The F1 score, which balances both precision and recall, was calculated at 87.50%, further emphasizing the model's capability to provide a robust performance in this critical domain.

| Metric | Proportion | Score |
| --- | --- | --- |
| Accuracy | 841 / 1000 | 84.10% |
| Precision | 550 / 653 | 84.23% |
| Recall | 550 / 606 | 90.76% |
| F1-Score | 0.8432 / 0.9076 | 87.50% |

Table 2: Metric Results

A closer inspection of the model's responses revealed specific patterns in its strengths and weaknesses. The LLM performed exceptionally well on straightforward mathematical problems, such as dosage calculations for medications and interpreting basic laboratory results, with accuracy rates reaching up to 90%. These tasks often utilized direct and clear numerical inputs, allowing the model to deliver correct outputs reliably. The ability to handle well-defined numerical tasks underscores the potential applicability of LLMs in routine healthcare decisions where rapid and accurate calculations are essential.

However, challenges surfaced when the tasks involved more complex numerical reasoning, particularly when interpreting advanced statistical data or handling multiple steps in a calculation process. In these instances, the accuracy dropped to about 75%, indicating a struggle with non-linear problem-solving and multi-step reasoning. The challenges associated with these tasks highlight the necessity for further refinement of LLM training methodologies, as the inherent complexity of healthcare data requires a higher level of interpretative skill.

Furthermore, verification methods applied during the evaluation process had a significant impact on the model's performance. When the fact-checking pipeline was utilized, the model's accuracy improved by approximately 11%, underscoring the importance of integrating verification processes into LLM applications in healthcare. Models equipped with real-time validation mechanisms were able to cross-reference generated outputs with verified results, thus reducing errors caused by incorrect numerical predictions and enhancing the overall reliability of LLM applications in critical healthcare settings.

The prompts engineered for the LLM also proved essential in navigating various numerical reasoning tasks. By refining the prompt structure and incorporating context-specific language, the model's performance improved significantly in healthcare scenarios where specificity and clarity in queries are paramount. The iterative process of prompt engineering revealed that well-structured queries could reduce ambiguity and facilitate better understanding by the model, aligning outputs more closely with desired healthcare standards.

Analysis of the empirical evidence suggests that while LLMs possess substantial potential to bolster numerical reasoning tasks in healthcare, challenges remain in ensuring that they can handle more intricate calculations and reasoning demands. The findings emphasize the need for ongoing research that delves deeper into the contextual needs of healthcare practitioners, particularly addressing how LLMs can be tailored to manage the complexities of clinical decision-making environments.

In summary, the findings from this study affirm the promise of LLMs in enhancing numerical reasoning capabilities within healthcare applications. They underscore the need for further explorations into optimizing model performance through refining training methodologies, improving prompt design, and integrating robust verification processes. The journey toward harnessing LLMs as indispensable tools in healthcare continues as researchers seek to overcome current limitations and expand the role of these advanced models in clinical practice.

## 4 Discussion

### 4.1 Key Contributions

The findings of this study highlight significant advancements in the application of large language models (LLMs) in healthcare, particularly in numerical reasoning tasks critical to clinical decision-making and patient care. The results demonstrate the ability of LLMs to generate accurate and contextually relevant outputs, with an overall accuracy of 84.10%, supported by balanced precision, recall, and F1-scores. These outcomes underscore the transformative potential of LLMs in automating routine numerical reasoning tasks, such as medication dosage calculations and health statistics assessments.

Key innovations, such as prompt engineering and a fact-checking pipeline, were instrumental in improving the model's reliability. Tailored prompts minimized ambiguity and ensured task-specific outputs, while the fact-checking mechanism cross-referenced outputs with verified data, reducing both false positives and false negatives. These enhancements not only improved the model's accuracy by 11% during validation but also addressed the critical issue of reliability in high-stakes healthcare applications.

**4.2 Strengths of the Approach**

The curated dataset used in this study provided a strong foundation for evaluating the model. By combining 500 real-world healthcare scenarios with 500 synthetic cases designed to challenge the model's numerical reasoning capabilities, the dataset ensured a robust and comprehensive assessment. The inclusion of detailed annotations with step-by-step reasoning pathways enabled a deeper analysis of the model's strengths and weaknesses.

Prompt engineering emerged as a key driver of performance, aligning with the findings of Jiang (2024), who emphasized the importance of structured prompts for enhancing LLM outputs. This study demonstrated how carefully designed prompts can refine outputs to meet clinical needs, reducing the cognitive load on healthcare professionals.

The verification pipeline further strengthened the approach, ensuring the reliability of numerical outputs. Inspired by Roch (2024), this post-hoc fact-checking mechanism addressed one of the most critical challenges of LLM deployment: the risk of erroneous outputs in sensitive contexts like healthcare.

**4.3 Challenges and Limitations**
While the study achieved promising results, several challenges and limitations were identified.

**4.3.1 Data Diversity**
The dataset, although robust, may not fully capture the complexity and diversity of real-world clinical scenarios. For example, rare diseases, multi-comorbid conditions, and interdisciplinary healthcare cases were underrepresented.

**4.3.2 Mathematical and Contextual Errors**
The model occasionally struggled with multi-step numerical calculations, where accuracy dropped to 75% in specific scenarios. This highlights the need for further refinement of the model's numerical reasoning capabilities.

**4.3.3 Integration Barriers**
Resistance from healthcare professionals accustomed to traditional workflows presents a potential barrier to the real-world implementation of LLMs. Furthermore, ethical concerns such as over-reliance on AI and lack of interpretability in model outputs require careful consideration.

**4.34 Resource Demands**

The computational resources required for training and deploying the modified GPT-3 architecture may limit its scalability, particularly in resource-constrained settings.

### 4.4 Implications for Healthcare

The study demonstrates the potential of LLMs to revolutionize healthcare workflows by automating routine numerical reasoning tasks, thereby reducing clinician workload and improving efficiency. By supporting clinicians in tasks such as treatment planning and resource allocation, LLMs can enable more informed and timely decision-making. However, for these technologies to gain widespread adoption, robust verification mechanisms, interpretability frameworks, and clinician trust must be prioritized.

The integration of LLMs into clinical workflows requires balancing their capabilities with the need for human oversight. Real-time fact-checking pipelines and tailored training datasets can ensure outputs remain accurate and contextually appropriate, mitigating risks associated with AI in healthcare.

### 4.5 Future Directions

#### 4.5.1 Dataset Expansion

Expanding the dataset used for training and evaluation is critical to improving the generalizability of LLMs in healthcare. Collaborating with healthcare institutions can facilitate the creation of diverse and representative datasets that reflect real-world clinical complexity. This includes incorporating rare disease cases that are often underrepresented in existing datasets, as well as multimodal data such as medical images, laboratory reports, and patient histories. For example, pairing textual clinical notes with corresponding diagnostic images or lab values could enable LLMs to learn correlations between different data types, enhancing their predictive capabilities. Furthermore, creating annotated datasets that include rare edge cases, such as uncommon presentations of diseases or complex comorbidities, can ensure that the models perform robustly across a wider range of clinical scenarios. This effort would require standardized data collection and annotation practices to ensure quality, privacy compliance, and consistency.

#### 4.5.2 Model Refinement

Addressing the limitations in numerical reasoning and complex multi-step calculations requires refining the architecture of LLMs. One promising approach is the exploration of hybrid architectures that integrate LLMs with numerical reasoning engines or symbolic computation frameworks. For example, pairing the LLM's linguistic capabilities with a symbolic computation tool, such as WolframAlpha or Python's sympy, could allow for precise handling of complex mathematical operations while maintaining the language-based reasoning strengths of the model. Additionally, incorporating attention mechanisms that prioritize numerical reasoning steps, or training specialized modules dedicated to arithmetic tasks, could further enhance performance. Refinements should also focus on improving the model's ability to explain its numerical reasoning process, ensuring that outputs are both accurate and interpretable. These hybrid solutions could address the current gaps in handling intricate calculations while preserving the LLM's versatility in processing diverse healthcare inputs.

### 4.5.3 Interpretability and Trust

The lack of transparency in LLM outputs is a significant barrier to their adoption in clinical settings. Developing explainable AI (XAI) frameworks can help bridge this gap by providing clinicians with clear and actionable insights into how the model arrived at its predictions. Techniques such as SHAP (SHapley Additive exPlanations) or LIME (Local Interpretable Model-agnostic Explanations) can be used to identify which inputs were most influential in generating specific outputs, allowing clinicians to verify the reasoning process. For numerical reasoning tasks, models could also generate step-by-step explanations for calculations, making it easier for healthcare professionals to validate the results. Building trust further requires comprehensive testing of LLMs across diverse clinical scenarios to ensure consistent performance. Engaging clinicians in the development and validation process can foster confidence in the technology, ensuring it aligns with real-world clinical workflows and decision-making standards.

### 4.5.4 Scalability

The scalability of LLMs in healthcare is constrained by their computational requirements, making it challenging to deploy them in resource-constrained environments, such as rural clinics or low-income regions. To address this, lightweight architectures should be developed that maintain strong performance while requiring fewer computational resources. Techniques such as model pruning, knowledge distillation, or quantization could reduce model size and inference latency without sacrificing accuracy. Additionally, leveraging cloud-based solutions can enable centralized processing of LLMs, allowing healthcare facilities with limited infrastructure to access advanced AI capabilities through remote systems. However, cloud-based strategies must address concerns about data privacy and latency, particularly in time-sensitive healthcare applications. Ensuring scalability also involves cost optimization, such as reducing energy consumption during training and inference, to make these solutions accessible to a broader range of healthcare providers.

### 4.5.5 Longitudinal Studies

While the immediate performance of LLMs in controlled testing scenarios is promising, longitudinal studies are essential to evaluate their effectiveness in real-world clinical workflows over extended periods. These studies should focus on assessing the impact of LLM integration on key outcomes, such as patient care quality, diagnostic accuracy, treatment efficiency, and clinician satisfaction. For example, tracking the use of LLMs in automating numerical reasoning tasks over time can reveal trends in error rates, workflow improvements, or potential areas of concern. Longitudinal research can also identify unintended consequences, such as over-reliance on AI systems or resistance from clinicians due to lack of familiarity or trust. Additionally, these studies can help refine the models by identifying scenarios where performance deviates from expectations, allowing for iterative improvements. Conducting such research across diverse healthcare environments, including low-resource settings and specialized clinics, can provide insights into the adaptability and scalability of LLMs in different contexts. Ultimately, longitudinal studies will be instrumental in building the evidence base needed to justify widespread adoption of LLMs in healthcare.

### 4.6 Conclusion

The findings of this study underscore the transformative potential of LLMs in healthcare, particularly for numerical reasoning tasks essential to clinical decision-making. By leveraging techniques such as prompt engineering and verification pipelines, LLMs can significantly improve the accuracy and reliability of automated healthcare solutions. Addressing the challenges of data diversity, integration, and computational demands is essential to fully realize this potential. With ongoing research and targeted advancements, LLMs can be effectively integrated into healthcare systems, paving the way for improved patient care and operational efficiency.